\documentclass[a4paper]{article}

\usepackage{amsmath}
\usepackage{mathtools}
\usepackage{amsthm}
\usepackage{amsfonts}
\usepackage{bbm}
\usepackage{graphicx}
\usepackage[dvipsnames]{xcolor}
\usepackage{array}
\usepackage[hyphens]{url}

\usepackage{tikz}
\usetikzlibrary{shapes.geometric, arrows.meta, positioning}

\usepackage[pdftex,bookmarksnumbered,bookmarksopen,colorlinks,citecolor={blue!70!black},linkcolor={blue!70!black},urlcolor=blue]{hyperref}
\usepackage[capitalize]{cleveref}
\usepackage[round]{natbib}
\usepackage{parskip}
\usepackage{fullpage}
\usepackage{libertine}
\usepackage{inconsolata}
\usepackage{authblk}
\renewcommand{\cite}{\citep}

\usepackage{hyperref}
\let\svthefootnote\thefootnote
\newcommand\freefootnote[1]{
  \let\thefootnote\relax
  \footnotetext{#1}
  \let\thefootnote\svthefootnote
}

\author[1]{Bogdan Kulynych$^*$}
\author[2]{Theresa Stadler$^*$}
\author[1,3]{Jean Louis Raisaro}
\author[4,5]{Carmela Troncoso}

\affil[1]{Lausanne University Hospital}
\affil[2]{Swiss Data Science Center, EPFL}
\affil[3]{University of Lausanne}
\affil[4]{Max Planck Institute for Security \& Privacy}
\affil[5]{EPFL}
\date{\footnotesize\today}

\title{Should I use Synthetic Data for That? An Analysis of the Suitability of Synthetic Data for Data Sharing and Augmentation}

\providecommand{\UsePackageFor}[2]{ \ifx#2\undefined\usepackage{#1}\fi }

\UsePackageFor{amssymb}{\boxtimes}
\usepackage[normalem]{ulem}		
\usepackage{bbding}				
\usepackage{endnotes}			
\usepackage{hyperref}			
\usepackage{resource/comment/hyperendnotes}	
\usepackage{pdfcolfoot}			
\usepackage{xcolor}

\ifdefined\OrigFootnote\relax\else
	\newenvironment{FootnoteContent}{}{}
	\let\OrigFootnote\footnote
	\let\OrigFootnoteText\footnotetext
	\renewcommand{\footnotetext}[1]{\OrigFootnoteText{\begin{FootnoteContent}#1\end{FootnoteContent}}}
	\renewcommand{\footnote    }[1]{\OrigFootnote    {\begin{FootnoteContent}#1\end{FootnoteContent}}}
\fi

\definecolor{PurplePlum}{rgb}{0.1,0,0.55}
\definecolor{Brown}{rgb}{0.5,.25,0}
\definecolor{Orange}{rgb}{1,.3,0}
\definecolor{Gray}{rgb}{.7,.7,.7}
\definecolor{DarkGreen}{rgb}{.1,.41,.1}
\definecolor{Turquoise}{HTML}{00CED1}

\newif\ifBleck
\newcommand\Bleck {\Blecktrue} 

\newcommand\Colour[1] {\color{#1}}

\newcommand\PrintToCLinks{	
  {\Colour{blue}\mbox{
    \hyperlink{w1619}{\sf$\rightarrow$~top}\quad
    \hyperlink{w1031}{\sf$\rightarrow$~toc}\quad
    \hyperlink{w1148}{\sf$\rightarrow$~lof}\quad
    \hyperlink{GreenRoom}{\sf$\rightarrow$~gr}\quad
    \hyperlink{EndNotes}{\sf$\rightarrow$~en}\quad
    \hyperlink{Sargasso}{\sf$\rightarrow$~sg}\quad
    \hyperlink{Index}{\sf$\rightarrow$~idx}
  }}
}
\makeatletter 
\newcommand\ToCLinks{
  \ifx\@onlypreamble\@notprerr		
    \hypertarget{w1619}{}			
  \else
    \AtBeginDocument{\hypertarget{w1619}{}}	
  \fi

  \ifBleck\else	
    \ifdefined\cofoot
      \cofoot{\PrintToCLinks}
      \cefoot{\PrintToCLinks}
    \else
      \def\@oddfoot{\PrintToCLinks}
      \def\@evenfoot{\PrintToCLinks}
    \fi
 \fi
}
\makeatother

\newif\ifEndNotes 

\newcommand\FnSym{{\scriptsize\PencilLeftDown\kern.1em}}		
\newcommand\EnSym {{$\bigtriangledown$}}

\def\MarkupsHowto{} 
\newcommand{\MarkupsHowtoAdd}[1]{\expandafter\def\expandafter\MarkupsHowto\expandafter{\MarkupsHowto{}#1}} 

\newif\ifMarkupsHowtoPrinted 
\newif\ifSuppress 

\newcommand\MakeMarkups[3][.]{

     \Suppressfalse
     \ifBleck\Suppresstrue\fi
     \ifx0#1\Suppresstrue\fi
     \ifx1#1\Suppressfalse\fi

     \expandafter\providecommand\csname#2x\endcsname {} 
     \ifSuppress\expandafter\renewcommand\csname#2x\endcsname{\relax}\else
                       \expandafter\renewcommand\csname#2x\endcsname{#3}\fi

     \expandafter\providecommand\csname#2\endcsname {} 
     \ifSuppress\expandafter\renewcommand\csname#2\endcsname[1]{##1}\else
                       \expandafter\renewcommand\csname#2\endcsname[1]{{\csname#2x\endcsname##1}}\fi

     \expandafter\providecommand\csname#2d\endcsname {} 
     \ifSuppress\expandafter\renewcommand\csname#2d\endcsname[1]{\relax}\else
                       \expandafter\renewcommand\csname#2d\endcsname[1]{{\csname#2x\endcsname\sout{##1}}}\fi

     \expandafter\providecommand\csname#2r\endcsname {} 
     \ifSuppress\expandafter\renewcommand\csname#2r\endcsname[2]{{##2}}\else
                       \expandafter\renewcommand\csname#2r\endcsname[2]{\csname#2d\endcsname{##1} \csname#2\endcsname{##2}}\fi

     \expandafter\providecommand\csname#2i\endcsname {} 
     \ifSuppress\expandafter\renewcommand\csname#2i\endcsname[1]{\relax}\else
                       \expandafter\renewcommand\csname#2i\endcsname[1]{\csname#2\endcsname{##1}}\fi

     \expandafter\providecommand\csname#2t\endcsname {} 
     \ifSuppress\expandafter\renewcommand\csname#2t\endcsname[1]{\relax}\else
                       \expandafter\renewcommand\csname#2t\endcsname[1]{{\csname#2x\endcsname{\mbox{$\langle\!\langle$}##1{\csname#2x\endcsname\mbox{$\rangle\!\rangle$}}}}}\fi

     \expandafter\providecommand\csname#2b\endcsname {} 
     \ifSuppress\expandafter\renewcommand\csname#2b\endcsname[1][empty]{\relax}\else
                       \expandafter\renewcommand\csname#2b\endcsname[1][\empty]{\ifx\empty##1\empty
                       	\label{#2-bookmark} 
                              \marginpar [\raggedleft\csname#2\endcsname{{\footnotesize\fbox{#2 working here}}~$\Longrightarrow$}]
                                                {\csname#2\endcsname{$\Longleftarrow$~{\footnotesize\fbox{#2 working here}}}}
                       \else 
                       	\marginpar [\raggedleft\csname#2\endcsname{\ifx\empty##1\empty\else\fbox{\tiny\parbox{8em}{\raggedright##1}}~\fi$\Longrightarrow$}]
                                                {\csname#2\endcsname{$\Longleftarrow$\ifx\empty##1\empty\else~{\tiny\fbox{\parbox{8em}{\raggedright##1}}}\fi}}\fi}\fi

     \expandafter\providecommand\csname#2TD\endcsname {} 
     \ifSuppress\expandafter\renewcommand\csname#2TD\endcsname{\relax}\else
                       \expandafter\renewcommand\csname#2TD\endcsname{\csname#2\endcsname{\fbox{#2 to do}}}\fi

     \expandafter\providecommand\csname#2Bar\endcsname {} 
     \ifSuppress\expandafter\renewcommand\csname#2Bar\endcsname{\relax}\else
                       \expandafter\renewcommand\csname#2Bar\endcsname{\csname#2\endcsname{\scriptsize\XSolidBrush}}\fi

     \expandafter\providecommand\csname#2f\endcsname {} 
     \ifSuppress\expandafter\renewcommand\csname#2f\endcsname[2][]{\relax}\else
      \expandafter\renewcommand\csname#2f\endcsname[2][\empty]{ 
        {\mbox{\csname#2x\endcsname\tiny$\boxtimes$}\marginpar{\hsize1cm\csname#2x\endcsname\fbox{\FnSym\footnotemark}}\relax
        \footnotetext{\csname#2x\endcsname##2}}}\fi

     \expandafter\providecommand\csname#2e\endcsname {}
     \ifSuppress\expandafter\renewcommand\csname#2e\endcsname[1]{\relax}\else%
      \expandafter\renewcommand\csname#2e\endcsname[1]{%
       \global\EndNotestrue
       \mbox{\scriptsize\csname#2x\endcsname$\boxtimes$}\relax%
       \marginpar{\hsize1cm\csname#2x\endcsname\fbox{\EnSym\endnotemark%
                          \hypertarget{ENmark\thepage-\theendnote}{}~\hyperlink{ENtext\thepage-\theendnote}{{\Colour{blue}$\downarrow$}}}%
       }%
       {
        \def\zz{\noexpand#3}%
        \edef\z{~{[Endnote \theendnote\ %
        on p.\noexpand\hypertarget{ENtext\thepage-\theendnote}{}\thepage%
                    ~\noexpand\hyperlink{ENmark\thepage-\theendnote}{{\noexpand\Colour{blue}$\uparrow$}}]}%
        }%
        \expandafter\endnotetext\expandafter{\z\vspace{2ex}\\ ##1\newpage}%
       }
      }\fi

     \expandafter\providecommand\csname#2n\endcsname {}
     \ifSuppress\expandafter\renewcommand\csname#2n\endcsname[1]{\relax}\else%
      \expandafter\renewcommand\csname#2n\endcsname[1]{%
       \global\EndNotestrue
    \marginpar{{\tiny\endnotemark}\hypertarget{ENmark\thepage-\theendnote}{}~\hyperlink{ENtext\thepage-\theendnote}{}}
       {
        \def\zz{\noexpand#3}%
        \edef\z{~{\zz[Endnote (deferred) 
        from p.\noexpand\hypertarget{ENtext\thepage-\theendnote}{}\thepage%
        ]}%
        }%
        \expandafter\endnotetext\expandafter{\z\vspace{2ex}\\ ##1\newpage}%
       }
      }\fi

     \expandafter\providecommand\csname#2fe\endcsname {} 
     \ifSuppress\expandafter\renewcommand\csname#2fe\endcsname[2][]{\relax}\else 
      \expandafter\renewcommand\csname#2fe\endcsname[2][]{ 
       \def\File{##1}\relax
       \ifx\File\empty\csname#2f\endcsname{##2}\else 
        \global\EndNotestrue 
        \mbox{\scriptsize\csname#2x\endcsname$\boxtimes$}
        \marginpar{\csname#2x\endcsname\fbox{\FnSym\footnotemark}}\relax
        \footnotetext{~\csname#2x\endcsname##2\
                             --- See [\EnSym\endnotemark\hypertarget{ENmark\thepage-\theendnote}{}
                             \kern-.2em\hyperlink{ENtext\thepage-\theendnote}{{\Colour{blue}$\downarrow$}}].}\relax
       { 
         \def\zz{\noexpand#3}
         \edef\z{~{\zz[Endnote~\thefootnote~on~p.\noexpand\hypertarget{ENtext\thepage-\theendnote}{}\thepage
                     ~\noexpand\hyperlink{ENmark\thepage-\theendnote}
                     {{\noexpand\Colour{blue}\kern-0.1em$\uparrow$}]}}
                     {\noexpand\footnotesize\noexpand\newline\noexpand\hspace*{2em} (~from file {\noexpand\tt\File.tex}~)}
         }
         \expandafter\endnotetext\expandafter{\z~\par\input{##1}\newpage}
        } 
       \fi 
      } 
     \fi 

     \ifSuppress\relax\else\ifBleck\relax\else
      \MarkupsHowtoAdd{\par\csname#2t\endcsname{
       $\backslash$\texttt{#2}$\cdots$\ markups are in \textbf{this} colour\ifx#1..\else\ifx1#1.\else, e.g.\ for #1.\fi\fi
       \ifMarkupsHowtoPrinted\relax\else 
        \global\MarkupsHowtoPrintedtrue 
        \begin{quote}\begin{tabular}{l@{\hspace{2em}}p{.7\linewidth}}
         \multicolumn{2}{l}{\texttt{$\backslash$MakeMarkups\ifx#1.\relax\else[#1]\fi\{#2\}\{{\it$\langle$colour command\/$\rangle$}\}}
         				 --- Defines the macros below:}\\
             & see comments at \texttt{$\backslash$MakeMarkups} definition. \\[1ex]
         \texttt{$\backslash$#2\{$\langle$text$\rangle$\}} & Sets \texttt{$\langle$text$\rangle$} in \texttt{#2}'s colour. \\
         \texttt{$\backslash$#2x} & Changes to \texttt{#2}'s colour (until end of context). \\
         \texttt{$\backslash$#2d\{$\langle$text$\rangle$\}} & Sets \texttt{$\langle$text$\rangle$} in \texttt{#2}'s colour with a strikethrough (i.e.\ delete). \\
         \texttt{$\backslash$#2r\{$\langle$this$\rangle$\}\{$\langle$that$\rangle$\}} &
          Strikes through \texttt{$\langle$this$\rangle$} and inserts \texttt{$\langle$that$\rangle$} (i.e.\ replace). \\
         \texttt{$\backslash$#2f\{$\langle$text$\rangle$\}} & Meta-comment: puts \texttt{$\langle$text$\rangle$} in a \texttt{#2}-footnote with a {\tiny$\boxtimes$} in the main text. \\
         \texttt{$\backslash$#2t\{$\langle$text$\rangle$\}} & Use for meta when  \texttt{$\backslash$#2f} isn't allowed (``Not in outer-par mode.'') \\
         \texttt{$\backslash$#2b[$\langle$optional$\rangle$]} & Marginal pointer, with label for hyper-linking directly there. \\
         \texttt{$\backslash$#2e\{$\langle$text$\rangle$\}} & Puts \texttt{$\langle$text$\rangle$} in a \texttt{#2}-endnote with a (big) $\boxtimes$ in the main text. \\[.5ex]
         \texttt{$\backslash$#2n\{$\langle$text$\rangle$\}} & Like \texttt{$\backslash$#2e}
         except there's no reference from the main text. Good for ``decluttering''
         when you still want to have the footnote- or endnote texts as reminders. \\[.5ex]
         \texttt{$\backslash$#2fe[$\langle$this$\rangle$]\{$\langle$that$\rangle$\}} & Makes a \texttt{$\backslash$#2f\{$\langle$that$\rangle$\}} that refers to a \\
           & \texttt{$\backslash$#2e\{$\langle$contents of file this.tex$\rangle$\}}. \\
           & Without the optional argument, acts as \texttt{$\backslash$#2f\{$\langle$that$\rangle$\}}. \\[.5ex]
         \texttt{$\backslash$#2Bar} & Inserts ``burn after reading'' symbol \csname#2Bar\endcsname, meaning
          \begin{quote}\begin{itemize}\setlength\itemsep{0pt}
           \item If yours is the only \csname#2Bar\endcsname\ in this (presumably someone else's) footnote, and you are happy that the footnote has been addressed,
           go ahead and comment-out the whole footnote. (The \csname#2Bar\endcsname\ is their request for you to ``approve and remove''.)
           \item If you are not happy, delete only your \csname#2Bar\endcsname\ and follow-on in the footnote
            (in your colour, i.e.\ with \texttt{$\backslash$#2x}) saying why you are not happy.
           \item If you are happy, but there are others' burn-after-reading symbols as well as yours, just delete yours; the other people have not yet responded.
          \end{itemize}
          \end{quote}
          The idea is that when everyone's happy, the last person will comment-out the meta-text. \\[0.5ex]
         \texttt{$\backslash$#2TD} & Inserts {\csname#2TD\endcsname}\ . \\
        \end{tabular}\end{quote}
       \fi
      }}
     \fi\fi
}

\newif\ifNoGreenRoom
\newcommand\MakeGreenRoom {\ifBleck\relax\else\ifNoGreenRoom\relax\else
\newcommand\NewGRLabel[1] {\OldGRLabel{GreenRoom-##1}} 
 \newcommand\NewGRRef[1] 
 {\expandafter\ifx\csname r@GreenRoom-##1\endcsname\relax\OldGRRef{##1}\else\OldGRRef{GreenRoom-##1}\fi}
 \let\OldGRLabel\label \let\label\NewGRLabel
 \let\OldGRRef\ref \let\ref\NewGRRef
 \hrule
 ~\\\begin{center}\Huge \hypertarget{GreenRoom}{Green Room}
 \end{center}~\\
 \hrule
\fi\fi}

\newcommand\EndGreenRoom  {\ifBleck\relax\else\ifNoGreenRoom\relax\else
\let\label\OldGRLabel
\let\ref\OldGRRef
\fi\fi}

\newif\ifNoEndNotes

\newif\ifNoSargasso
\newcommand\MakeSargasso {
 \hypertarget{Sargasso}{}
 \newcommand\NewLabel[1] {\OldLabel{Sargasso-##1}} 
 \newcommand\NewRef[1] 
 {\expandafter\ifx\csname r@Sargasso-##1\endcsname\relax\OldRef{##1}\else\OldRef{Sargasso-##1}\fi}
 \let\OldLabel\label \let\label\NewLabel
 \let\OldRef\ref \let\ref\NewRef
\ifBleck\end{document}\else\ifNoSargasso
\relax
\else
  \hrule
  ~\\\begin{center}\Huge Sargasso
  \end{center}~\\
  \hrule
 \fi\fi
}

\newcommand\EndSargasso  {\ifBleck\relax\else\ifNoSargasso\relax\else
\let\label\OldLabel
\let\ref\OldRef
\fi\fi}

\newcommand\EndDocument {\ifBleck\end{document}\fi} 

\newcommand\Cite[2][\empty] {{\Colour{red}\ifx#1\empty[#2]\else[#2,~#1]\fi}}

\MakeMarkups[Theresa]{T}{\Colour{PurplePlum}}
\MakeMarkups[Bogdan]{B}{\Colour{Turquoise}}

\newtheorem{problem}{Problem Formalisation}

\newcommand{\sR}{\mathbb{R}}

\newcommand{\sD}{\mathbb{D}}
\newcommand{\dataSource}{S}
\newcommand{\dataSyn}{\smash{\tilde{S}}}
\newcommand{\augSource}{S_\mathsf{aug}}
\newcommand{\dataBase}{S_\mathsf{base}}

\newcommand{\trueParam}{\theta}
\newcommand{\params}{\theta}
\newcommand{\train}{\mathtt{train}}
\newcommand{\genModel}{G}
\newcommand{\Loss}{L}

\newcommand{\targetFunc}{f}
\newcommand{\queryFamily}{\mathcal{F}}
\newcommand{\base}{\mathsf{base}}
\newcommand{\aug}{\mathsf{aug}}

\DeclareMathOperator{\E}{\mathbb{E}}

\newcommand{\para}[1]{\vspace{1mm}\noindent\textbf{#1.}}

\Bleck

\begin{document}

\maketitle

\freefootnote{$^*$ Contributed equally}

\begin{abstract}
    \noindent
    Recent advances in generative modelling have led many to see synthetic data as the go-to solution for a range of problems around data access, scarcity, and under-representation. In this paper, we study three prominent use cases: (1) Sharing synthetic data as a proxy for proprietary datasets to enable statistical analyses while protecting privacy, (2) Augmenting machine learning training sets with synthetic data to improve model performance, and (3) Augmenting datasets with synthetic data to reduce variance in statistical estimation.
    For each use case, we formalise the problem setting and study, through formal analysis and case studies, under which conditions synthetic data can achieve its intended objectives.
    We identify fundamental and practical limits that constrain when synthetic data can serve as an effective solution for a particular problem. Our analysis reveals that due to these limits many existing or envisioned use cases of synthetic data are a poor problem fit.
    Our formalisations and classification of synthetic data use cases enable decision makers to assess whether synthetic data is a suitable approach for their specific data availability problem.
\end{abstract}

\section{Introduction}

Synthetic data is frequently seen as a potentially transformative solution to critical challenges in data-driven research and innovation~\citep{jordon2022synthetic,van2023beyond,liu2024best,kapania2025examining}.
Its proponents claim that synthetic data can serve as a drop-in replacement for real data that protects privacy but preserves its full analytical value~\citep{van2023beyond, JRC2022, MostlyAISharing, MDClone, JRC2024Financial, IEEESA}, can enhance machine learning training sets to improve the performance of downstream models~\citep{Bing22, van2023beyond, ghalebikesabi2023differentially, Juwara24, ktena2024generative, IEEESA}, and can augment small datasets to reduce variance of statistical estimates~\cite{liu2025synthetic,ElKabaji25}.
\emph{In this paper, we aim to understand if and when synthetic data is actually a suitable technical solution for these problems.}
Understanding the fundamental and practical limitations of synthetic data is critical to avoid the downsides of inappropriate uses. Developing generative models and creating synthetic datasets can require substantial infrastructural investment or fees paid to solution providers. Moreover, synthetic data applied in unsuitable problem settings can lead to biased models or invalid and misleading conclusions~\citep{whitney2024real, wyllie2024fairness}. In high-stakes applications, such as clinical trial research, development of diagnostic models in healthcare or risk models in welfare and finance, the potential biases and errors introduced by synthetic data could cause harm to individual's health or incur financial and economic costs~\citep{wilkinson2020time,FCA25}.

Borrowing a metaphor from \citet{narayanan2025ai}, asking whether ``synthetic data'' is a suitable tool for a given data-availability problem is akin to asking whether a ``vehicle'' is an appropriate solution to a given transportation problem. Synthetic data and its potential use cases are diverse and require special consideration in each case. In this paper, we introduce formalisations to disambiguate and understand synthetic data use cases based on three distinct objectives: (1) Sharing synthetic data as a proxy for replicating analyses on proprietary datasets, (2) Augmenting training data to improve model performance, and (3) Augmenting datasets to reduce variance in statistical estimation.

For each use case, we formalise the problem setting, identify fundamental and practical limits,
and examine case studies that demonstrate both successful and problematic applications. In our analysis, we focus on problem settings where statistical validity, privacy, or utility are required. Although synthetic data could serve other purposes, such as, testing software~\cite{Gray1994}, evaluations of models based on simulated data~\cite{baumann2023bias,liu2024best}, or exploratory analyses~\cite[see, e.g.,][]{kennedy2024planning}, we focus on high-stakes applications where the promise of synthetic data is to be ``as good as real data''~\cite{van2023beyond}. Overall, our work provides practitioners with concrete guidance to determine whether synthetic data is suitable for their applications.

\section{Sharing synthetic data as a proxy for replicating analyses on proprietary data}
\label{sec:sharing-proxy}

Widened data sharing and data publishing are considered essential drivers for future innovation and growth~\citep{EU_DataStrategy20, EU_DataReuse23, EU_DataSpaces22, FDS}. How to make high-value privacy-sensitive datasets, such as electronic health records~\citep{EHDS} or financial transaction data~\citep{FCA25}, available to a broader group of data users for analysis while preserving the privacy of affected individuals has become one of the key challenges of the data-driven economy~\citep{stadler2025purpose}. Sharing the real, proprietary data is often challenging due to privacy concerns, data protection restrictions, or other compliance regulations. In response, advocates propose to share synthetic in place of real data to enable the desired data uses and frequently advertise synthetic data as a ``better than real'' data replacement that preserves the full value of the proprietary data, i.e., that allows external entities to draw valid statistical inferences about the real from the synthetic data~\citep{BellovinDR19, JRC2022, van2023beyond, MostlyAI, MDClone}.

\subsection{Problem formalisation}
\label{subsec:formalisation_sharing}
In the synthetic data sharing setting, a \textit{data controller} holds a proprietary \textit{source dataset} $\dataSource \in \sD^n$, for $n > 0$, over the space of data records $\sD$. A \textit{data user} would like to run an \textit{analysis function} $\targetFunc(\cdot) \in \queryFamily_\mathsf{use}$ belonging to some family of analyses $\queryFamily_\mathsf{use}$ to learn information from the data $\targetFunc(\dataSource)$.
The function $\targetFunc(\cdot)$ could represent, for instance, a machine learning training algorithm or a statistical query function~\citep{FCA25, JRC2022}. Another type of analysis commonly cited in the relevant literature on synthetic data in the tabular domain are \emph{$k$-way marginal queries}~\cite[see, e.g.,][]{ponomareva2025dp}.
In a dataset describing demographics of people, including their gender, city of birth, and a year of birth, an example of a 3-way marginal query $\targetFunc(\dataSource)$ is the proportion of individuals in the dataset who are men born in Paris, France, in 1993.

In some data-sharing scenarios, the family of analysis functions $\queryFamily_\mathsf{use}$, or potentially even the specific function $\targetFunc(\cdot)$ that will be run by the data user, is well-defined and known to the data controller at the time of data generation. We refer to these cases as \textit{narrow-purpose data sharing}. Examples of narrow-purpose data sharing are synthetic data for the release of population statistics such as contingency tables~\citep{Shlomi25} or for learning the weights of a well-specified fraud detection model that leverages specific patterns in fraudulent transactions~\citep{FCA25}. In other cases, the analysis that will be run on the synthetic data is unknown to both the data controller and data user when sharing the data. Yet, the data user requires statistical validity for their eventual inferences. Example uses that fall in this category are the re-use of health data for secondary purposes through data spaces~\citep{stadler2025purpose, EHDS, TEHDAS}. We refer to these cases as \textit{broad-purpose data sharing}.

\begin{figure}[h]
    \centering
    \includegraphics[width=\linewidth]{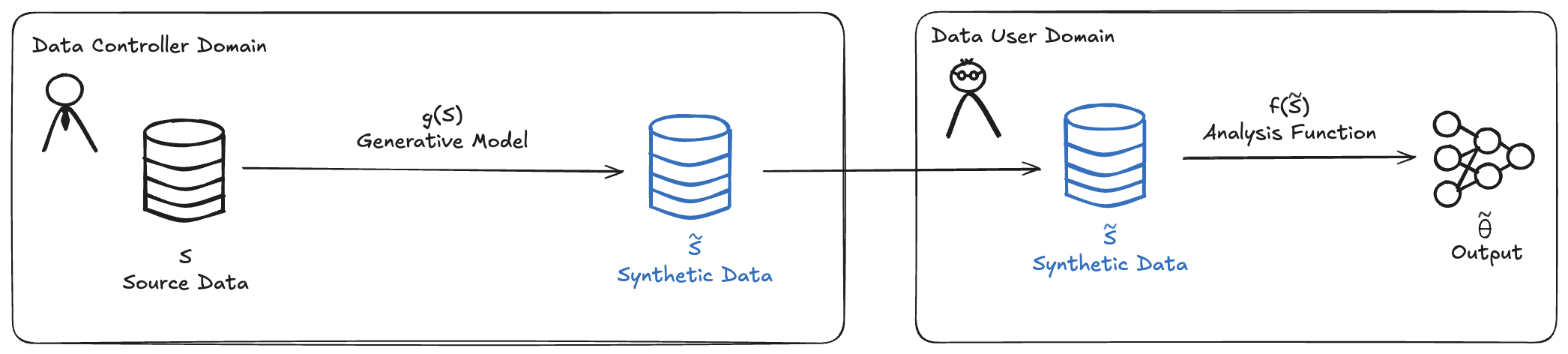}
    \caption{\textbf{Sharing synthetic data as a proxy for replicating analyses on proprietary data:} To overcome data access barriers, the data controller shares a synthetic in place of the real data with the data user who runs the desired analysis $\targetFunc(\dataSyn)$ on the synthetic instead of the real source data $\dataSource$.}
    \label{fig:sharing}
\end{figure}

The entity that holds the source data $\dataSource$, the data controller, and the entity that defines and runs the analysis function $\targetFunc(\cdot)$, the data user, \textit{are two separate entities}. In some cases, these could be two fully separate real-world entities, like a hospital and a pharmaceutical company in a clinical-trial setting, in others the data controller and the data user could be represented by two teams within the same organisation.
What all data sharing problems have in common is one underlying challenge: Due to privacy concerns or other compliance issues, the data controller is unable or is unwilling to share the real source data with the data user for the intended purpose. In this paper, we specifically focus on the salient case where privacy is the main barrier to data sharing.
The idea of using synthetic data as a solution to this problem is relatively simple. The data controller generates a \emph{synthetic dataset} $\dataSyn \sim \genModel(\dataSource)$ using a \emph{generative model} trained on the real source data $\dataSource$, denoted by the random variable $\genModel(\dataSource)$ over $\sD^n$. Then, the data controller shares $\dataSyn$ in place of $\dataSource$ with the data user. The data user computes $\targetFunc(\dataSyn)$ treating the synthetic data as a stand-in proxy for the real source data. We illustrate this setup in \cref{fig:sharing}, where the intended use is training a machine learning model.

To fulfil its objective in this setting, the synthetic data should thus satisfy two requirements: (1) \emph{replicate statistical patterns from the real source data} such that inferences drawn by the data user are the same as if they had run the analysis on the real data, i.e., provide statistical validity, and (2) ensure that realistic adversaries \emph{cannot reconstruct information about individuals} in the proprietary dataset, i.e., provide privacy.
We formalise this objective in the following way:
\vspace{.5em}
\begin{problem}[Sharing synthetic data as a proxy for replicating analyses]
    In the problem of \emph{sharing synthetic data as a proxy}, a data controller wants to find a generative model $\genModel(S)$ that achieves the following goals:
    \begin{enumerate}
        \item \emph{Validity.} Ensure that a synthetic dataset $\dataSyn$ sampled from a generative model $\genModel(\dataSource)$ trained on $\dataSource$ ensures validity for a class of analyses $\queryFamily_\mathsf{use}$ that the data user is interested in:
        \begin{equation}
            \label{eq:validity}
            \targetFunc(\dataSource) \approx \targetFunc(\dataSyn), \; \forall\targetFunc \in \queryFamily_\mathsf{use}
        \end{equation}
     where $\approx$ denotes the closeness of outputs with high probability over the randomness of $\genModel(\dataSource)$. This definition is also known as \emph{accuracy of queries}~\cite{dwork2014algorithmic}.
        \item \emph{Privacy.} For any attack within a relevant threat model denoted as $A(\dataSyn)$ and individual target record $x \in \sD$, the attack cannot distinguish between synthetic datasets generated with or without the said target record:
        \begin{equation}
            \label{eq:privacy}
            A(\genModel(\dataSource \setminus \{x\})) \approx A(\genModel(\dataSource \cup \{x\})),
        \end{equation}
        where $\approx$ denotes similarity in distribution. This is a simplified version of the standard definition of differential privacy~\cite{dwork2014algorithmic}. Although \cref{eq:privacy} is framed in terms of membership disclosure, it also immediately implies that fully or partially reconstructing records is challenging~\cite[see, e.g.,][]{kulynych2025unifying}.
    \end{enumerate}
\end{problem}

In the remainder of the section, we identify certain conditions under which achieving validity and privacy simultaneously is impossible. Such impossibility---when it arises---does not imply that synthetic data is  unsuitable as a data proxy in principle. Rather, it means that its scope is limited to relatively low-stakes applications where validity for analysis functions run on the synthetic data is not a strict requirement. An example of such application could be running exploratory analyses~\cite{kennedy2024planning} where the goal is to generate plausible hypotheses. In such cases, it is important to ensure transparency by communicating clearly to all entities that synthetic data does not guarantee validity of the results.

\subsection{Synthetic data as a drop-in replacement for real data lacks statistical validity}
A crucial issue with synthetic data shared as a proxy for proprietary data is that synthetic data generated through common generative methods, such as GANs~\cite{Goodfellow2014} or diffusion models~\cite{ho2020denoising}, \emph{cannot ensure validity for statistical analyses}. The main appeal of using synthetic data to enable statistical analysis, compared to other possible technical solutions such as interactive query answering systems~\citep[see, e.g.,][]{aymon2024lomas} or federated learning algorithms~\citep{mcmahan2017communication}, is that synthetic data has the same format as real data. It is therefore tempting for the data user to simply run any statistical analysis they are interested in on the synthetic dataset and treat the output as if it had come directly from the real proprietary data. Unfortunately, this approach does not guarantee statistical validity for the outcomes of the analyses because, in general, it is unknown, to both the data controller and the data user, which properties of the source data the synthetic data replicates. The findings and outcomes of statistical analyses made on synthetic data, are in an epistemic limbo: they could be indicative of signal in the original data, but there is usually no way of knowing how reliable a finding is.

\para{Empirical approaches}
A standard approach to manage this limitation is for the data controller to conduct \emph{replication studies}: Empirically evaluate whether certain findings made on a proprietary dataset replicate on the synthetic data prior to sharing. The data controller chooses a test set of analysis functions $\queryFamily_\mathsf{test}$ and computes the discrepancies $|\targetFunc(\dataSource) - \targetFunc(\dataSyn)|$ for all $\targetFunc \in \queryFamily_\mathsf{test}.$ For example, a common practice is to evaluate differences between all $2$-way marginals in the synthetic and proprietary source data~\cite{kaabachi2025scoping}.
This approach can indicate how well a synthetic data generation method performs with respect to the set of analysis functions tested $\queryFamily_\mathsf{test}$, e.g., in the case of 2-way marginals---how well the synthetic data preserves correlations.
Unfortunately, this approach is unlikely to solve the validity problem in practice, as explained next.

One way to ensure the validity of findings derived by the data user from the synthetic data through $\targetFunc(\dataSyn)$ for some function $\targetFunc \in \queryFamily_\mathsf{use}$, is to set $\queryFamily_\mathsf{test} = \queryFamily_\mathsf{use}$. In most cases, this is not a feasible option and an unrealistic assumption that largely defeats the purpose of synthetic data as a proxy~\citep{stadler2025purpose}: If both the data controller and data user know the specific quantities of interest $\queryFamily_\mathsf{use}$ in advance and the data controller can compute $\targetFunc(\dataSource)$ for all $\targetFunc \in \queryFamily_\mathsf{use}$ to compute distances, the data controller could simply share the intended analysis outcomes, e.g., model weights or summary statistics, instead of the synthetic data.
Within the broad framework of testing, one possibility to ensure validity is to derive formal bounds on the validity of $\targetFunc(\dataSyn)$ for $\targetFunc \in \queryFamily_\mathsf{use}$ even when $\queryFamily_\mathsf{test} \neq \queryFamily_\mathsf{use}$. Indeed, in some cases, it is possible to bound $|\targetFunc(\dataSyn) - \targetFunc(\dataSource)|$ based on the empirically tested quantities  $|\targetFunc'(\dataSyn) - \targetFunc'(\dataSource)|$ for some $\targetFunc' \notin \queryFamily_\mathsf{use}$. For example, it is possible to bound the difference in the loss of logistic regression trained on synthetic data and on the source data using the difference in low-order $k$-way marginals between synthetic and source data~\cite{zhou2024bounding}. Deriving such bounds for more complex analyses, however, is non-trivial.

\para{Tailored generation} Another approach to address the validity problem is to construct synthetic datasets that, by design, are guaranteed to preserve validity for pre-defined fixed families of analyses $\queryFamily_\mathsf{use}$~\cite{raghunathan2003multiple,mckenna2021winning,Shlomi25}.
The challenge with this approach is that existing methods for doing so are only available for quite simple statistical analyses such as low-order $k$-way marginals~\cite{ponomareva2025dp}. As mentioned previously, general-purpose approaches to create synthetic data, e.g., models such as diffusion models, largely lack the capacity to ensure statistical validity for specific analysis functions.

\subsection{Synthetic data cannot ensure validity for too many analyses and preserve privacy at the same time}

A related issue is a fundamental theoretical limitation known as the \emph{fundamental law of information recovery}~\citep{dwork2014algorithmic}. In a nutshell, this principle states that any synthetic dataset (or any other sharing mechanism such as an interactive query system) inevitably compromises privacy if it is guaranteed to provide sufficiently accurate answers for too many questions about the source data.
The formal foundation for this limitation was established through the theory of reconstruction attacks, first developed in a seminal work by \citet{dinur2003revealing}. This theory shows that an attacker that obtains approximately correct answers to a sufficiently large collection of simple statistical queries about a dataset can reconstruct the records of nearly every individual in that dataset.
For instance, if an attacker can obtain accurate enough answers to just $O(n)$ queries on a dataset of $n$ individuals, they can reconstruct the secrets bits of the majority of records in the dataset.
This is a blatant privacy violation, and violates the requirement in \cref{eq:privacy}.

As a result of this law, if a synthetic dataset is \emph{guaranteed} to provide valid inferences across a set of analysis functions $\queryFamily_\mathsf{use}$---such as all $k$-way marginals for some $k > 1$, or all linear combinations of features,---an adversary could apply reconstruction techniques to recover private information from the original dataset~\cite{dwork2017exposed}. The synthetic nature of the records itself does not protect against such attacks~\cite{stadler2022synthetic, Ganev25}.

Despite these theoretical foundations having been established more than two decades ago, it is common to find synthetic data generation approaches that assume that generating plausible-looking fake records is sufficient for privacy protection~\cite[see, e.g.,][]{kaabachi2025scoping}. Under the fundamental law of information recovery, the reality is that there exists a hard limit on the number and complexity of statistical analyses $\queryFamily_\mathsf{use}$ that can be accurately preserved while simultaneously maintaining meaningful privacy guarantees. Attempting to preserve validity of too many analyses, even the simplest ones such as pairwise correlations between features ($2$-way marginals), can result in blatant privacy violations.

Differential privacy provides a rigorous framework to navigate the trade-off between privacy and validity~\cite{dwork2014algorithmic}. Differentially private algorithms can indeed generate synthetic data that provably protects individual privacy while preserving certain statistical properties~\cite[see, e.g.,][for a review]{ponomareva2025dp}. Differential privacy, however, does not circumvent the limits imposed by the fundamental law of information recovery. Instead, it provides tools to explicitly choose the trade-off between privacy and validity.

\subsection{Case studies}

\para{Differentially private release of the 2020 US Census microdata}
The United States Census Bureau provides a case study of privacy-preserving synthetic data sharing deployed at a massive scale. Motivated by reconstruction attacks on data products from the 2010 census that demonstrated that it was possible to re-identify individuals and reconstruct their census responses with high accuracy, the Census Bureau adopted formal differential privacy methods for the 2020 decennial census~\citep{abowd2018us}. Rather than releasing public use microdata files or traditional tabular summaries, the Bureau now generates privacy-protected statistics in the format of synthetic census microdata using sophisticated algorithms that add carefully calibrated noise~\citep{abowd20222020, stanley2024synthetic, ruggles2025shortcomings}.

The algorithm used by the Census Bureau preserves many key statistics about demographic distributions, geographic patterns, and population characteristics. These enable specific, well-defined important uses of census data, such as population counts for geographic redistricting or demographic breakdowns for resource allocation~\citep{abowd20222020}. Although the techniques have met substantial criticism~\cite{ruggles2019differential,pujol2020fair, ruggles2025shortcomings}, e.g., due to disparate impact in apportionment, the U.S. census case provides a blueprint for the use of synthetic data as a potential solution to privacy-preserving data sharing. The generated synthetic dataset transparently takes into account both statistical validity for a carefully curated set of queries, and privacy, with provable guarantees on both aspects. The approach enables the users of synthetic data to be relatively certain about the results of statistical analyses so long as they are within the supported set.

\para{Sharing health data for secondary data uses in the European Health Data Space}
One of the most prominent goals of the European Health Data Space (EHDS) regulation is to facilitate cross-border health data sharing for secondary data uses~\citep{EHDS, TEHDAS}. The intended uses of EHDS data span an exceptionally wide range of applications: epidemiological research, health technology assessment, public health surveillance, pharmaceutical development, and health system planning, among others. Many of these use cases by definition require statistical validity for a great range of analysis functions, involve complex multivariate analyses, temporal patterns, or causal inference, and assume that the specific analysis functions run by data end users are not defined at the time of data sharing~\citep{TEHDASutility}.
The EHDS puts substantial pressure on healthcare providers to make data more widely available for secondary uses while complying with strict privacy regulations. In response, synthetic data solution providers have suggested that sharing synthetic in place of real patient data could achieve these goals~\citep{IEEESA, MDClone, MostlyAISharing}. Proponents argue that ``synthetic data have the same utility and can be analysed in the same way as the original, real-world dataset — all the while preserving patient privacy''~\citep{MDClone}, or make equivalent claims~\cite{MostlyAISharing}. As we discussed previously, however, synthetic data \emph{cannot} reliably support all intended data uses without either compromising privacy or sacrificing validity due to the well-established fundamental law of information recovery.
The discussions surrounding the use of synthetic data to implement data sharing for secondary purposes under the EHDS exemplify the fundamental challenges that make synthetic data unsuitable as a technical solution to support \emph{broad-purpose} data sharing initiatives where the validity of outcomes could be important.

\subsection{Takeaways}

\para{Principled privacy-preserving data sharing requires a pre-specified and narrow purpose}
The fundamental law of information recovery establishes that no algorithm can simultaneously guarantee statistical validity for arbitrary analyses and preserve privacy. Therefore, synthetic data is a suitable approach to privacy-preserving data sharing only if the intended family of analyses is narrow and well-defined at the time of data generation. When the data sharing purpose is genuinely broad but validity is required, alternative technical solutions such as secure computation environments or restricted data access with governance controls are necessary.

\para{Validity guarantees require specialized generative models}
When the data sharing purpose is narrow with a well-defined family of analyses, it is possible to generate synthetic data with validity guarantees through specialized generation methods tailored to the target analyses. For instance, when all intended analyses can be derived from $k$-way marginals, well-established methods such as MST~\cite{mckenna2021winning} and AIM~\cite{mckenna2022aim} can provide both validity and privacy guarantees. Generic generative models such as diffusion models generally lack such guarantees. Extending the available toolbox to support more complex analysis families with validity guarantees is an important direction for future research.

\section{Augmenting training datasets with synthetic data}
\label{sec:aug-ml}

Data scarcity is a substantial issue in real-world deployments of ML models~\citep{Alzubaidi2023, Bing22, Juwara24, liu2014}. Models trained on insufficient data often fail to generalise to new domains~\citep{Gianfrancesco2018, liu2014}, with potentially harmful consequences~\citep{zech2018, Dexter2020, Bing22}.
Addressing data scarcity is a challenging problem, as collecting more data is often infeasible~\citep{Hripcsak11, Kaplan14, Alzubaidi2023}. Traditional imbalance-correction methods that oversample existing records~\citep{Idrissi22} or use linear interpolation~\citep{Chawla02} have shown some improvements but might fail for datasets with strong biases or complex minority distributions~\citep{chen2023, Juwara24}.
More recently, \emph{synthetic data augmentation} (SDA) in which data created with a generative model is used to augment training sets has emerged as an alternative solution. SDA has been studied across multiple data modalities~\citep{Antoniou17, Jain22, Fonseca23}, application domains~\citep{Dina22,Das22,Bing22}, and bias-related issues~\citep{Antoniou17, Das22, Dina22, Juwara24, Bing22}.
SDA is claimed to improve model performance by expanding limited training sets. In particular, it is claimed to enable the generation of ``more representative data sets that lead to fairer downstream models''~\citep{Bing22}, to improve performance ``for underrepresented groups''~\citep{van2023beyond}, and to be ``a robust solution to mitigate data bias''~\citep{Juwara24}.
In cases of under-representation, however, recent work highlights that these efforts risk producing mere ``caricatures'' of minority groups rather than authentic representation~\cite{kapania2025examining}.
In this section, we introduce a formal definition of data augmentation problems for machine learning model training, and discuss conditions under which synthetic data can and cannot achieve the intended objectives.

\subsection{Problem formalisation}
In the augmentation setting, a \emph{model developer} holds a \emph{base dataset} $\dataBase \sim P^n$, for $n > 0$ and some distribution $P$ over the space of data records $\sD$. The model developer trains a machine learning model using their base dataset $\dataBase$ in order to obtain \emph{model parameters} $\params_\base = \train(\dataBase)$. To improve the performance, the model developer would like to augment the available base data with synthetic data $\dataSyn$. The synthetic dataset is sampled from a generative model, $\dataSyn \sim \genModel(\augSource)$ that is based on an \emph{augmentation source} $\augSource$. The augmentation source $\augSource$ could represent the base dataset $\dataBase$ itself, an external dataset, domain knowledge, e.g., a causal model, or a pre-trained model, e.g., an off-the-shelf language model. Thus, the augmentation source could be based on proprietary information directly available to the model developer, on public data, or on data from another external party. We illustrate these various settings in \cref{fig:augmentation} (A) and (B).
We denote the machine learning model trained on the augmented dataset, i.e., merged data from both $\dataBase$ and $\dataSyn$ as $\params_\aug = \train(\dataBase, \dataSyn)$.

Once the model developer has completed training, a \emph{model operator} deploys the trained model. In some augmentation settings, the model developer also deploys the model and simply seeks to improve performance on their own data through augmentation~\citep{Juwara24, liu2024best}, depicted in \cref{fig:augmentation} (C). In other settings, the model developer and operator could be two separate entities. For example, a model developer, Hospital A, could have developed a clinical prediction model, and a model operator, Hospital B, could be interested in deploying the model on their patient population~\cite{wilkinson2020time, ktena2024generative} which we illustrate in \cref{fig:augmentation} (D).

To measure the performance of a model $\params$, we assume the existence of a target distribution $Q$, and a function $\Loss_Q(\params)$ that measures the error of the model.
For instance, the function $\Loss_Q(\params)$ could represent the expected error of the model on the test distribution $\Loss_Q(\params) = \E_{z \sim Q}[\ell(\params; z)],$
for some loss function $\ell(\params; z) \geq 0$.

\begin{problem}[Synthetic data augmentation of training data]
	In the \emph{synthetic data augmentation for training data problem}, the model developer that holds the base dataset $\dataBase$ wants to find a generative model $\genModel(\cdot)$ that generates synthetic data $\dataSyn \sim \genModel(\augSource)$ such that augmenting the training data of a machine learning model with the synthetic data decreases the error on a target distribution $Q$ compared to the base model:
 	\begin{equation}\label{eq:aug_objective}
 	    \Loss_Q(\params_\aug) < \Loss_Q(\params_\base).
 	\end{equation}
\end{problem}

This setting is fundamentally different from sharing synthetic data as a proxy described in \cref{sec:sharing-proxy}. In synthetic data sharing, the objective is to \emph{replicate} certain findings from the proprietary source dataset on the synthetic data. In augmentation, the synthetic data is not expected to replicate the base dataset and there is no analysis done on the synthetic data per se. Therefore, there is no concern about the validity of analysis run on the synthetic data.

\subsection{To improve target performance, the augmentation source must contain relevant information}

A fundamental requirement for synthetic data augmentation to provide benefits is that the augmentation source $\augSource$ must contain information relevant to the target distribution $Q$ on which the model operator will evaluate the model. If the augmentation source and target distribution are misaligned, we cannot expect synthetic data generated from $\augSource$ to improve performance over using only the base dataset $\dataBase$.

\begin{figure}
    \centering
    \includegraphics[width=\linewidth]{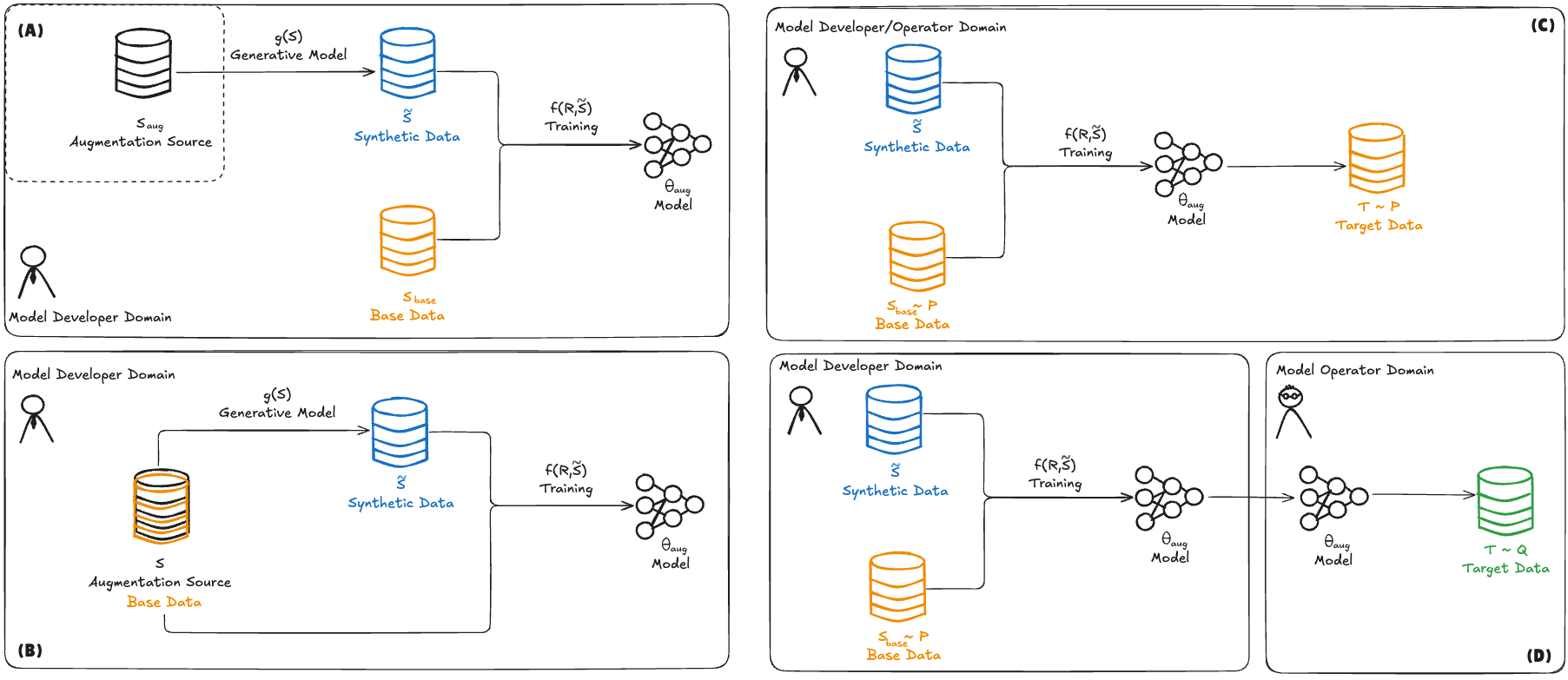}
    \caption{\textbf{Synthetic data augmentation problems can be categorised along two main dimensions.} The first dimension is the source information: (A) External augmentation source vs. (B) Bootstrapped augmentation. The synthetic data $\dataSyn$ used to augment the base data $\dataBase$ can either be (A) derived from an additional augmentation source or (B) derived from the base data itself. The second dimension is the goal: (C) Augmentation to increase sample size (in-distribution) vs. (D) Augmentation for domain generalisation (out-of-distribution). The goal of data augmentation can be to improve performance on (C) in-distribution samples such as minority groups from the base domain or (D) an unknown target domain that differs from the data available to the model developer.}
    \label{fig:augmentation}
    \vspace{-10pt}
\end{figure}

\para{Bootstrapped augmentation} A salient case~\citep{Antoniou17, Das22, Juwara24} is synthetic data generated directly from the base dataset itself, i.e., $\augSource = \dataBase$, shown in \cref{fig:augmentation} (B). This approach is known as \emph{bootstrapped augmentation}. The generative model $\genModel(\dataBase)$ learns patterns present in $\dataBase$ and produces synthetic samples $\dataSyn$ that reflect these patterns. If the base dataset $\dataBase \sim P^n$ is sampled from a distribution $P$ that systematically differs from the target distribution $Q$, the synthetic data will inherit and potentially amplify these differences.

For instance, suppose the base dataset comes from a distribution $P$ that exhibits systematic class imbalance or lacks representation of certain subgroups that are present in the target distribution $Q$. A generative model trained on $\dataBase$ could learn to reproduce the biased distribution and generate synthetic samples that mirror the existing biases rather than correcting them. Although bootstrapped augmentation might provide regularisation benefits by smoothing decision boundaries, it cannot add new information about the target distribution. In such cases, augmenting $\dataBase$ with synthetic data generated from itself cannot systematically address distribution shift without additional external information, though it might provide regularisation benefits in certain settings. Recent studies~\cite{manousakas2023usefulness,wahler2024evaluating} have shown that bootstrapped augmentation provides comparable or at most modest improvements over simple approaches such as random oversampling~\cite{Idrissi22} and SMOTE~\cite{Chawla02}. Thus, claims that \emph{purely bootstrapped} synthetic data augmentation, without additional source information, can generate ``more representative datasets'' or improve fairness for under-represented groups~\citep{Bing22, van2023beyond} likely overstate the reach of regularisation when the base data does not cover the target distribution.

\para{External augmentation sources} An alternative approach, shown in \cref{fig:augmentation} (A), is to use an external augmentation source $\augSource \neq \dataBase$, such as a dataset from a different institution, a pre-trained generative model, or a large-scale foundation model. External augmentation can lead to the desired benefits if $\augSource$ contains information about target $Q$ that is not already captured in $\dataBase \sim P^n$ .
The external source $\augSource$ can potentially improve generalisation to the target distribution if it provides complementary information. For instance, in a setting of augmenting the training data of a clinical prediction model, we could expect improvements if the external source is a dataset that contains examples from demographic groups or clinical presentations absent in $P$ but present in the target distribution $Q$.

Establishing whether the augmentation source provides complementary information requires careful analysis of both the augmentation source and target distribution.
Therefore, in augmentation settings where the synthetic data development and deployment settings systematically differ, e.g., sharing across hospitals as described above~\citep{ktena2024generative,wilkinson2020time}, we should not expect synthetic data to achieve its objective.

\subsection{Circular dependency of representative validation data}
Even when the augmentation source $\augSource$ contains relevant information about the target distribution $Q$, the model developer and operator face a critical challenge: How can they determine whether synthetic data augmentation improves model performance on the target distribution?

To empirically evaluate whether augmentation reduces the target error in \cref{eq:aug_objective}, i.e., whether $\Loss_Q(\params_\aug) < \Loss_Q(\params_\base)$, the model operator must have access to a test dataset that is sufficiently representative of $Q$, and large enough to provide reliable estimates of the loss function. For instance, if $L_Q(\cdot)$ is defined as expected loss $L_Q(\params) = \E_{z \sim Q}[\ell(\params; z)]$, it is normally approximated with a test dataset $T \sim Q^m$ for some $m > 1$, as $$
    L_Q(\params) \approx \smash{\frac{1}{m}}\sum_{z \in T} \ell(\params; z).
$$
Without such a test set, there is no empirical basis to determine whether the augmentation strategy has succeeded or failed. The model developer could assess the performance of a model trained on the augmented data $\theta_\aug$ on a validation set drawn from the same distribution as $\dataBase \sim P^n$, i.e., evaluate $L_Q(\params) \approx \E_{z \sim P}[\ell(\params; z)]$. However, this provides no useful information about generalization to $Q$ in cases where $P$ and $Q$ are systematically different. Thus, there might be no way to reliably evaluate the improvement in performance due to augmentation. In fact, this issue is not specific to synthetic data augmentation but a fundamental problem in the setting of under-represented or biased data.

Consider a concrete scenario depicted in \cref{fig:augmentation} (D): A hospital develops a clinical prediction model and augments its limited dataset with synthetic examples generated from a pre-trained generative model. Internal cross-validation shows improved performance with augmentation. If the hospital later deploys this model \emph{in a different clinical setting} with \emph{a different patient population}, however, the improvement observed might not materialize or could even reverse. Without access to representative data from the deployment setting during development, the hospital has no way to validate that its augmentation strategy is appropriate for external deployment. As a result, without a reliable method to assess performance, synthetic data augmentation risks negatively biasing the model and potentially hurting performance.

\subsection{Case studies}

\para{Diffusion models to improve fairness of diagnostic models under distribution shifts}
When certain demographic groups or clinical conditions are not well represented in their training data, the resulting models often exhibit lower accuracy for these subpopulations~\citep{Dexter2020, chen2021, Machado2022, wilkinson2020time}. Acquiring additional labelled examples for those groups could be impractical due to resource constraints or the inherent scarcity of relevant cases~\citep{Levine2020, ktena2024generative}.
Thus, a standard application of synthetic data augmentation is to ensure that a model developed on one population maintains performance when deployed on a different target population.
Recent work by \citeauthor{ktena2024generative} considers diffusion-based generative models, trained on external image collections that include unlabelled samples from the deployment setting, to produce targeted synthetic examples for groups that are poorly represented in the labelled training data. By specifying both diagnostic categories and demographic characteristics as conditioning variables, practitioners can direct the generative process to create training examples that fill specific gaps in the data. In experiments,
\citeauthor{ktena2024generative} find that mixing real and synthetic training samples can yield classifiers with improved consistency across demographic subgroups and better generalisation to new clinical sites.
The key to the success of this approach is \emph{external augmentation} which leverages additional information sources that genuinely contain relevant external information on the target distribution $Q$. Indeed, pre-trained models encode distributional knowledge from diverse populations and imaging conditions. Unlabelled data from the target domain, even without expert annotations, provides information about the deployment setting. This additional information has the potential to augmenting samples that improve the target performance.

Although this approach worked in the academic setting, in real-world applications practitioners must still confront the evaluation challenge discussed earlier in this section: validating whether augmentation achieves its objective requires representative \emph{labelled} test data from the target distribution. Such data might not be available to the model developer who thus cannot judge whether the method has been successful or not. This uncertainty comes with the corresponding risk of biases introduced by the synthetic data.

\para{Re-captioning training data to improve prompt adherence in image generators}
Another common application of synthetic data augmentation is to improve model performance on established benchmarks. For instance, a persistent challenge for text-to-image generative models is their tendency to overlook or misinterpret certain elements of input prompts~\citep{betker2023improving, liu2024best}. The low quality of image captions in training datasets significantly contributes to this problem. Images scraped from the web are typically paired with brief, noisy, or tangentially related text descriptions that omit crucial visual details.
\citeauthor{betker2023improving} show that augmenting training sets with images re-captioned by a custom captioning model that generates richer textual descriptions might significantly reduce this issue. Experiments demonstrate that diffusion models trained on machine-generated captions exhibit substantially better prompt adherence than those trained on original web-scraped text. These improvements in generation quality through data augmentation can be easily measured through alignment metrics between generated images and their corresponding prompts on common image generation benchmarks~\citep{betker2023improving}.
The external information introduced by the bespoke captioning model is the key factor that makes this augmentation approach work. The captioning model is fine-tuned on curated examples of detailed image descriptions and thus encodes knowledge about visual attributes that lacks from the original training sets. This knowledge transfers to the image generator through the re-captioned training pairs.

The crucial factor that enables the model operator to validate the approach is the availability of established benchmark datasets such as Drawbench~\cite{saharia2022photorealistic}. These benchmarks provide the operator with test data from the target distribution of detailed user prompts. In contrast to the previous use case, where sufficient test data might not be available in practice, in this case practitioners can verify that re-captioning improves performance with respect to the target distribution $Q$.

\subsection{Takeaways}

\para{Effective augmentation requires external information sources beyond the base dataset}
Augmenting training sets with synthetic data cannot improve generalization to a target distribution unless the augmentation source contains complementary information about that target. Bootstrapped augmentation can provide regularisation benefits for in-distribution generalization but does not systematically address distribution or domain shifts without external information. In contrast, external augmentation sources such as datasets from different institutions, pre-trained models, or encoded domain knowledge about the target distribution are more likely to fulfil the objective.

\para{Validation data from the target distribution is essential to assess improvements}
A critical distinction exists between validation data requirements and training data requirements: practitioners might have sufficient target data for validation (e.g., hundreds of samples) but insufficient data for training (e.g., thousands of samples). In this common scenario, synthetic data augmentation can be empirically validated and is viable.
In contrast, if the setting lacks even minimal validation data from the target domain, it is impossible to reliably verify whether synthetic data augmentation improves or harms performance. In such cases, there is a problem of circular dependency: insufficient access to data from the target domain is often what motivates the use of data augmentation techniques. But without representative test data one cannot reliably verify whether data augmentation, or any other algorithmic intervention, improves or harms performance. Collecting at least a small representative validation set or benchmark from the target distribution is essential before deploying augmented models. Otherwise augmentation risks introducing not mitigating bias.

\section{Augmentation or imputation for statistical estimation}
\label{sec:aug-stats}

Beyond training machine learning models, synthetic data augmentation or partial imputation has been proposed to address data scarcity in statistical inference problems, from generating data annotations in social sciences~\cite{angelopoulos2023prediction,ziems2024can} to augmenting clinical trial data with generated control patients~\cite{ElKabaji25, pammi2025digital}. In these settings, researchers aim to estimate unknown population parameters such as population means, treatment effects, or other estimands from limited samples. The goal of data augmentation in these settings is to reduce the variance and uncertainty of these estimates. Although similar to the augmentation setting in \cref{sec:aug-ml}, it requires its own problem formalisation and analysis. In contrast to data augmentation for machine learning models, in statistical estimation problems, by definition, it is not possible to empirically test the accuracy of the outputs. Thus, different issues arise when using synthetic data in these settings.

\subsection{Problem formalisation}

In the statistical estimation setting, a \emph{researcher} holds a \emph{base dataset} $\dataBase$ of $n$ samples that represent ``gold standard measurements''~\citep{angelopoulos2023prediction} over a population of interest. For instance, in a clinical trial, $\dataBase$ could represent patients for whom real treatment outcomes have been measured. The researcher's objective is to estimate an unknown population parameter $\trueParam \in \sR$, such as an average treatment effect, using an estimator $\hat{\theta}(\dataBase)$ computed from the base data.
The challenge is that collecting samples in $\dataBase$ is often costly which results in small sample sizes and high variance in the estimates. To address this issue, the researcher wants to augment the base data with additional information from an augmentation source, such as, predictions from machine learning models trained on unlabeled data, historical records, or external databases~\citep{angelopoulos2023prediction, ElKabaji25}. The researcher uses the augmentation source to generate (semi-)synthetic data $\dataSyn$ and augments the base dataset with the synthetic samples. The augmented estimator $\hat{\theta}(\dataBase, \dataSyn)$ combines information from both the base data and the synthetic values. The objective is to reduce the mean squared error of the estimate compared to using only the base data:

\begin{problem}[Synthetic data augmentation for statistical estimation]
	In the \emph{synthetic data augmentation for statistical estimation problem}, a researcher that holds a base dataset $\dataBase$ of size $n$ with ``gold standard'' data samples wants to find a generative model $\genModel(\cdot)$ that generates synthetic data $\dataSyn$ and an estimator $\hat{\theta}(\dataBase, \dataSyn)$ such that augmentation reduces the estimation error:
    \vspace{-.5em}
	\begin{equation}\label{eq:stat_objective}
		\mathbb{E}[(\hat{\theta}(\dataBase, \dataSyn) - \trueParam)^2] < \mathbb{E}[(\hat{\theta}(\dataBase) - \trueParam)^2],
	\end{equation}
	where the expectation is taken over the randomness of the estimation procedures.
\end{problem}

As mentioned previously, this setting differs from augmentation for reducing model error discussed in \cref{sec:aug-ml}. In learning tasks, the model operator, in principle, can evaluate whether augmentation improves performance by testing on validation data sampled from the target distribution. In statistical estimation, the true parameter $\trueParam$ by definition is unknown. There is no test set available to empirically verify whether $\hat{\theta}(\dataBase, \dataSyn)$ is closer to $\trueParam$ than $\hat{\theta}(\dataBase)$. The researcher thus cannot directly measure whether synthetic data augmentation has been beneficial, or, conversely has biased the outcome.

\subsection{Standard estimation methods do not provide valid inferences when applied to augmented data}

A critical issue with synthetic data augmentation for statistical inference is that standard statistical estimators applied to augmented datasets in a naive way generally do not provide valid confidence intervals or hypothesis tests~\citep{angelopoulos2023prediction, angelopoulos2023ppi++}. If the researcher treats synthetic data $\dataSyn$ as if it were real data and applies conventional statistical methods to the combined dataset $(\dataBase, \dataSyn)$, the resulting conclusions might be invalid.

\para{Lack of validity with naive usage} The fundamental problem is that synthetic or imputed data introduces systematic bias, even if the synthetic data appears plausible. Standard inference procedures assume that all observations are genuine samples from the target population and do not account for the additional errors introduced by synthetic data. Consequently, treating synthetic samples as equivalent to real samples leads to confidence intervals with incorrect coverage rates and hypothesis tests with inflated Type I error rates~\cite[see, e.g.,][]{angelopoulos2023prediction,byun2025valid}. The resulting inferences have no validity guarantees and cannot provide reliable scientific evidence.

In the case of sharing synthetic data to replicate analyses on real data, discussed in \cref{sec:sharing-proxy}, there exist some types of analyses for which we can obtain validity through empirical tests. In the case of synthetic data augmentation for statistical estimation, likely the only way to ensure validity is to use specialized estimation procedures.
Classical frameworks such as multiple imputation~\cite{rubin1993statistical}, recent general frameworks such as prediction-power inference (PPI)~\citep{angelopoulos2023prediction,angelopoulos2023ppi++}, design-based supervised learning~\cite{egami2023using}, and confidence-driven inference~\cite{gligoric2025can}, or application-specific methods such as Procova~\cite{schuler2020increasing} for randomised clinical trials, provide estimation procedures for augmented datasets with validity guarantees. For example, the PPI framework assumes access to a small labelled dataset $\dataBase$, a potentially large unlabeled dataset, and predictions from a machine learning model to impute the missing labels, which can be thought as a form of synthetic data. By carefully combining the predictions with the labeled samples and correcting for the bias introduced by imputed labels, PPI constructs estimators with provably valid confidence intervals on the imputed dataset $(\dataBase, \dataSyn)$~\citep{angelopoulos2023prediction}.

\para{The applicability of specialized methods is limited} Although the existing methods for deriving guarantees represent important progress, their applicability is limited.
First, these methods are only available for \emph{specific types of estimators and augmentation procedures}. For instance, PPI provides guarantees for estimating means, quantiles, and parameters of certain regression models. Extending these results to more complex estimators---such as parameters from structural equation models or estimates from sophisticated causal inference procedures---requires substantial future work. Researchers cannot simply apply these techniques to arbitrary statistical analyses. Moreover, most of the methods above only apply to partially imputed data, and the study of guarantees with fully synthetic data has only started recently~\cite{byun2025valid}. Therefore, finding methods that ensure validity of inferences under synthetic data augmentation is a developing area of research.
Second, even when using valid inference procedures, there is \emph{no guarantee that augmentation will improve estimation accuracy}. Whether incorporating synthetic data actually reduces variance or moves the estimate closer to the true parameter value depends on the quality of the synthetic data. If the generative model introduces bias or fails to capture relevant features of the target distribution, augmentation is unlikely to bring benefit.

\subsection{Case studies}

\para{Synthetically augmented control arms in clinical trials}
Recruiting a sufficient number of participants remains a major obstacle for the successful completion of randomized clinical trials~\citep{ElKabaji25}. Challenges to patient recruitment can arise from strict inclusion criteria, ethical considerations, and high costs. When trials fall short of their recruitment targets, the statistical power of target measurements is low which might lead to inconclusive results or even trial abandonment. To address this problem, some researchers have started to explore the use of generative models to synthesize\footnote{External \emph{real data} of control patients used to augment or replace a randomized control arm in a treatment-only or an under-recruited trial is sometimes referred to as a ``synthetic control arm''. In this paper, we do not use this term, and always use ``synthetic'' to refer to synthetic data.} additional patient records that augment the available trial data~\citep{ElKabaji25, pammi2025digital}.

\citet{ElKabaji25} propose to generate synthetic ``digital twins'' of real patients from a generative model trained on the limited set of recruited patient measurements. These synthetic patients are included alongside real participants in statistical analyses to estimate treatment effects. Proponents argue that this approach can salvage underpowered trials and reduce the costs and time required for clinical research.

The fundamental issue with this application of synthetic data augmentation is that standard statistical inference procedures applied to datasets augmented with synthetic patients provide \emph{no guarantee of validity of the statistical inferences}, which is crucial to preserve the high quality of evidence that randomized control trials bring to knowledge production. The design and analysis of clinical trials employ rigorous protocols that specify primary endpoints, analysis populations, and methods for estimating treatment effects.
When synthetic patients are added to the analysis datasets, these protocols become theoretically meaningless.
Thus, the results of the analyses, e.g., confidence intervals around estimated treatment effects, might have coverage rates far from their nominal levels, and hypothesis tests might have inflated Type I error rates.

To mitigate these issues, \citeauthor{ElKabaji25} evaluate whether synthetic augmentation produces more accurate treatment effect estimates through replication studies. Unfortunately, outside of the lab setting---in an actual randomized clinical trial that has failed to meet recruitment targets---such validation is not possible. Researchers cannot know whether $\hat{\theta}(\dataBase, \dataSyn)$ is closer to the true treatment effect $\trueParam$ than $\hat{\theta}(\dataBase)$ because the true effect is unknown. As we have seen in \cref{sec:sharing-proxy}, this puts the results of the analyses in an epistemic limbo: the estimates could be more accurate or less accurate than those from the real data alone.

A reliable application of synthetic data to reduce estimation variance for randomized clinical trials without reducing the quality of evidence must use methods that account for the bias introduced by synthetic data in a principled way. For instance, \citet{poulet2025prediction} leverage PPI in combination with the imputation of clinical trial outcomes using a predictive model, finding improvements in statistical power. Unlike the approach of naive synthetic augmentation, the approach of \citet{poulet2025prediction} ensures validity of the estimated treatment effect despite the usage of imputed outcomes thanks to the usage of variants of PPI. Further research is needed to understand whether it is possible to derive equivalent guarantees with fully synthetic data augmentation, as opposed to only partial imputation, using, e.g., the estimators from \citet{byun2025valid}.

\subsection{Takeaways}

\para{Valid inference from synthetically augmented or imputed data requires specialized procedures}
Treating synthetic data as equivalent to real data and applying conventional statistical methods to augmented datasets generally produces confidence intervals with incorrect coverage and hypothesis tests with inflated error rates. Specialized procedures that explicitly account for error due to synthetic data, such as prediction-powered inference (PPI)~\cite{angelopoulos2023prediction,angelopoulos2023ppi++}, are necessary to obtain validity guarantees. Without such procedures, the resulting inferences cannot be trusted at the same level as the inferences that only use real data.

\para{Available methods support limited settings}
Current specialized methods~\cite{angelopoulos2023prediction,angelopoulos2023ppi++,gligoric2025can,byun2025valid} provide validity guarantees only for specific types of estimators and augmentation procedures. These methods support limited classes of estimators, and typically are only compatible with imputation of partially available data rather than fully synthetic augmentation. Before relying on synthetic data for statistical estimation in high-stakes applications, practitioners must verify that their setting satisfies these requirements. Extending these methods to broader settings is an important research direction.

\section{Concluding remarks}
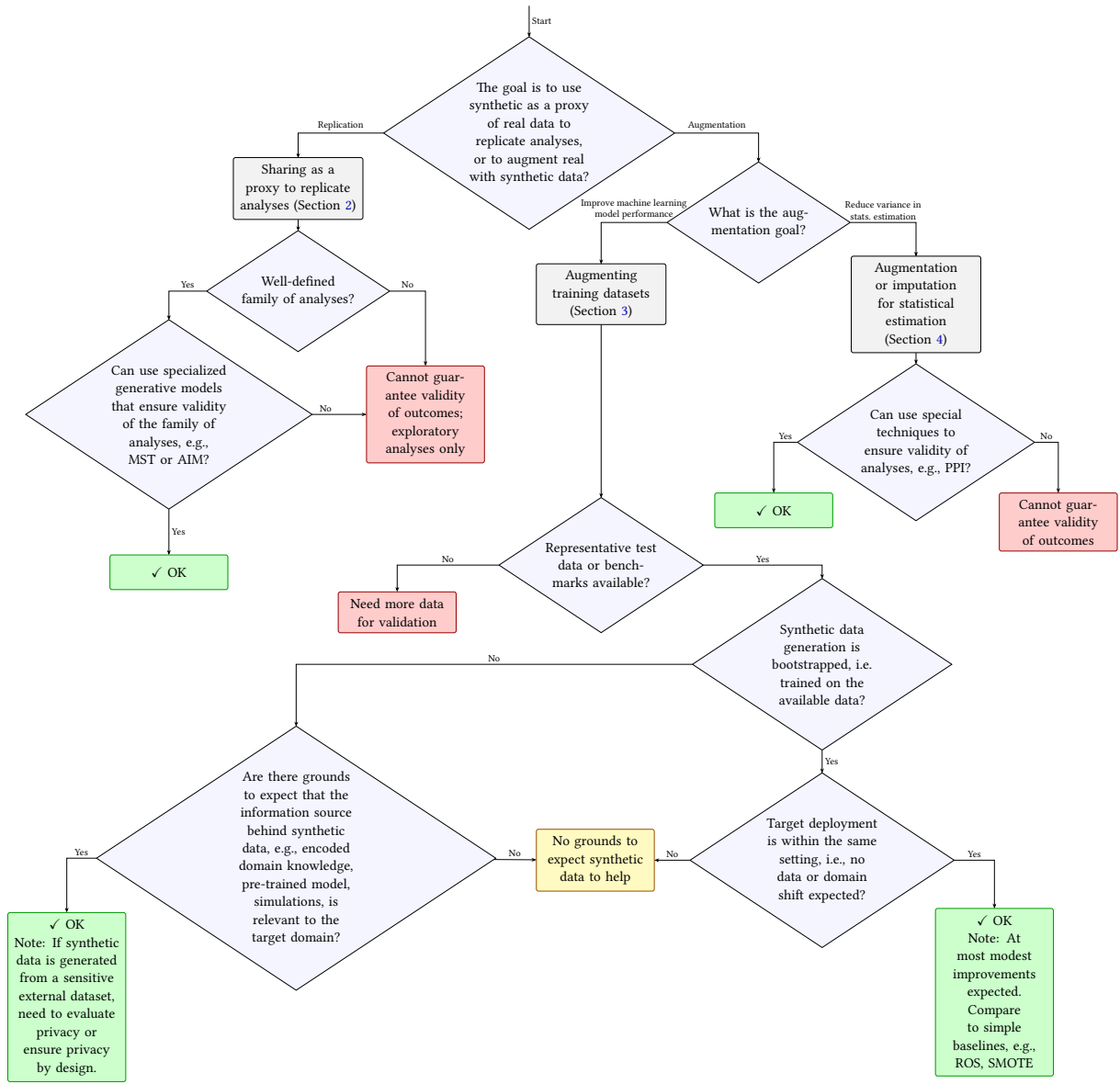
\begin{figure}[t!]
    \centering
    \tikzset{
  decision/.style={diamond, draw, fill=blue!4, text width=5.5cm, text centered, inner sep=2pt, minimum height=2cm, aspect=1.5, font=\LARGE},
  status/.style={rectangle, draw, fill=gray!10, text width=5cm, text centered, rounded corners, minimum height=1cm, inner sep=8pt, font=\LARGE},
  terminal-ok/.style={rectangle, draw=green!60!black, line width=1.5pt, fill=green!20, text width=4.5cm, text centered, rounded corners, minimum height=1.5cm, inner sep=8pt, font=\LARGE},
  terminal-warn/.style={rectangle, draw=orange!60!black, line width=1.5pt, fill=yellow!30, text width=4.5cm, text centered, rounded corners, minimum height=1.5cm, inner sep=8pt, font=\LARGE},
  terminal-stop/.style={rectangle, draw=red!60!black, line width=1.5pt, fill=red!20, text width=4.5cm, text centered, rounded corners, minimum height=1.5cm, inner sep=8pt, font=\LARGE},
  arrow/.style={thick,->,>=Stealth},
}

\resizebox{\linewidth}{!}{%
\begin{tikzpicture}[node distance=2.2cm and 3cm, scale=1, every node/.style={scale=1.5, font=\large}]

\node (start) [decision] {The goal is to use synthetic as a proxy of real data to replicate analyses, or to augment real with synthetic data?};

\node (entry) [above=2cm of start] {};

\node (sharing-status) [status, below left=of start, xshift=-2cm, yshift=2.5cm] {Sharing as a proxy to replicate analyses (\cref{sec:sharing-proxy})};
\node (sharing-q1) [decision, below=of sharing-status, yshift=1cm] {Well-defined family of analyses?};
\node (sharing-q2) [decision, below left=of sharing-q1, xshift=1.5cm, yshift=-0.5cm] {Can use specialized generative models that ensure validity of the family of analyses, e.g., MST or AIM?};
\node (sharing-ok) [terminal-ok, below=of sharing-q2, yshift=-0.48cm] {\checkmark~OK};
\node (sharing-exploratory) [terminal-stop, right=of sharing-q2, xshift=0.3cm, yshift=0cm] {Cannot guarantee validity of outcomes; exploratory analyses only};

\node (aug-q1) [decision, below right=of start, xshift=2.75cm, yshift=1cm] {What is the augmentation goal?};

\node (aug-ml-status) [status, below left=of aug-q1, xshift=0cm, yshift=1cm] {Augmenting training datasets (\cref{sec:aug-ml})};
\node (aug-ml-q1) [decision, below=of aug-ml-status, yshift=-6cm] {Representative test data or benchmarks available?};
\node (aug-ml-no-data) [terminal-stop, below left=of aug-ml-q1, xshift=-2cm, yshift=1.8cm] {Need more data for validation};
\node (aug-ml-q2) [decision, below right=of aug-ml-q1, xshift=2.5cm, yshift=0.5cm] {Synthetic data generation is bootstrapped, i.e. trained on the available data?};
\node (aug-ml-q3) [decision, below=of aug-ml-q2, yshift=.5cm] {Target deployment is within the same setting, i.e., no data or domain shift expected?};
\node (aug-ml-boot-no) [terminal-warn, left=of aug-ml-q3, xshift=0.5
cm, yshift=0cm] {No grounds to expect synthetic data to help};
\node (aug-ml-boot-yes) [terminal-ok, below right=of aug-ml-q3, xshift=0cm, yshift=1.3cm] {\checkmark~OK\\{Note: At most modest improvements expected. Compare to simple baselines, e.g., ROS, SMOTE}};
\node (aug-ml-q4) [decision, below left=of aug-ml-q2, xshift=-13.5cm, yshift=-2.2cm] {Are there grounds to expect that the information source behind synthetic data, e.g., encoded domain knowledge, pre-trained model, simulations, is relevant to the target domain?};
\node (aug-ml-ext-yes) [terminal-ok, below left=of aug-ml-q4, xshift=-1cm, yshift=2cm] {\checkmark~OK\\{Note: If synthetic data is generated from a sensitive external dataset, need to evaluate privacy or ensure privacy by design.}};

\node (aug-stats-status) [status, below right=of aug-q1, xshift=0cm, yshift=1.33cm] {Augmentation or imputation for statistical estimation (\cref{sec:aug-stats})};
\node (aug-stats-q1) [decision, below=of aug-stats-status, yshift=1cm] {Can use special techniques to ensure validity of analyses, e.g., PPI?};
\node (aug-stats-yes) [terminal-ok, below left=of aug-stats-q1, xshift=1cm, yshift=1cm] {\checkmark~OK};
\node (aug-stats-no) [terminal-stop, below right=of aug-stats-q1, xshift=-1cm, yshift=1cm] {Cannot guarantee validity of outcomes};

\draw [arrow] (entry) -- node[right] {Start} (start);
\draw [arrow] (start) -| node[near start, above] {Replication} (sharing-status);
\draw [arrow] (start) -| node[near start, above] {Augmentation} (aug-q1);

\draw [arrow] (sharing-status) -- (sharing-q1);
\draw [arrow] (sharing-q1) -| node[near start, above] {Yes} (sharing-q2);
\draw [arrow] (sharing-q1) -| node[near start, above] {No} (sharing-exploratory);
\draw [arrow] (sharing-q2) -- node[right] {Yes} (sharing-ok);
\draw [arrow] (sharing-q2) -- node[near start, above] {No} (sharing-exploratory.west);

\draw [arrow] (aug-q1) -| node[near start, above, align=center] {Improve machine learning\\model performance} (aug-ml-status);
\draw [arrow] (aug-q1) -| node[near start, above, align=center] {Reduce variance in\\stats. estimation} (aug-stats-status);

\draw [arrow] (aug-ml-status) -- (aug-ml-q1);
\draw [arrow] (aug-ml-q1) -| node[near start, above] {No} (aug-ml-no-data);
\draw [arrow] (aug-ml-q1) -| node[near start, above] {Yes} (aug-ml-q2);
\draw [arrow] (aug-ml-q2) -- node[right] {Yes} (aug-ml-q3);
\draw [arrow] (aug-ml-q3) -- node[midway, above] {No} (aug-ml-boot-no);
\draw [arrow] (aug-ml-q3) -| node[near start, above] {Yes} (aug-ml-boot-yes);
\draw [arrow] (aug-ml-q2) -| node[near start, above] {No} (aug-ml-q4);
\draw [arrow] (aug-ml-q4) -| node[near start, above] {Yes} (aug-ml-ext-yes);
\draw [arrow] (aug-ml-q4) -- node[midway, above] {No} (aug-ml-boot-no);

\draw [arrow] (aug-stats-status) -- (aug-stats-q1);
\draw [arrow] (aug-stats-q1) -| node[near start, above] {Yes} (aug-stats-yes);
\draw [arrow] (aug-stats-q1) -| node[near start, above] {No} (aug-stats-no);

\end{tikzpicture}
}
    \caption{Decision chart for potential uses and limits of synthetic data as a solution to data availability problems.}
    \label{fig:decision-chart}
\end{figure}

In this paper, we have introduced formalisations and classifications to understand when synthetic data is a suitable technical solution to problems of data access, scarcity, and under-representation. We studied three classes of applications of synthetic data: sharing as a proxy for replicating analyses on a proprietary dataset (\cref{sec:sharing-proxy}), augmentation to improve machine learning model performance (\cref{sec:aug-ml}), and augmentation to reduce variance in statistical estimation (\cref{sec:aug-stats}). Our analysis does not cover all potential applications. For instance, we do not consider simulation settings where synthetic data is used for model evaluation in controlled environments~\cite[see, e.g.,][]{baumann2023bias,liu2024best}. For each class of applications we do consider, we formalised the problem setting, identified critical limitations, and examined conditions under which synthetic data can or cannot achieve the intended objectives. Based on our results, we provide a summarising flow-chart for decision-making in \cref{fig:decision-chart}.

Across all three use cases, we identify fundamental and practical limitations that determine when synthetic data is a good problem-solution fit.
For sharing synthetic data as a proxy, the law of information recovery establishes that it is impossible to simultaneously preserve
privacy and ensure validity for arbitrary analyses. Synthetic data can only achieve these goals for well-defined sets of analyses specified at the time of sharing. It is unsuitable as a method for general-purpose data sharing if validity of statistical inferences is required.
For augmentation of machine learning training data, practitioners face two critical challenges: the augmentation source must contain complementary information about the target distribution beyond what the base dataset provides, and practitioners need representative test data from the target distribution to validate the augmentation objective.
For augmentation to reduce variance in statistical inference, standard methods applied naively to augmented or imputed data produce invalid inferences. Practitioners must therefore use specialized procedures such as prediction-powered inference that provide guarantees for specific estimators and settings.

Although these three use cases face different challenges, a common thread underlies their limitations: The problem of \emph{verifying or ensuring that synthetic data achieves its intended goal}.
In some cases, such as synthetic data augmentation for statistical estimation, it is impossible to empirically assess whether the intended objectives have been met.
In other cases, such as synthetic data augmentation for training machine-learning models, validation is possible if one can assume access to sufficient representative test data. The lack of access to data that would enable such validation, however, could be precisely what motivated the use of synthetic data in the first place.
When it is impossible to ensure or verify the outcomes, the results obtained through the use of synthetic data remain in an \emph{epistemic limbo}. Specialised methods for data generation or estimation can resolve the issue by providing formal guarantees. Such methods, however, are limited in scope and not always available in practice, with substantial future research required to solve this problem.

Our work discusses prominent use cases where synthetic data is proposed as a solution to data access, scarcity, and under-representation problems. We hope that our results provide practitioners with guidance to determine whether synthetic data constitutes an appropriate solution for their application and to identify challenges they might encounter.

\section*{Acknowledgements}
This project is partially supported by Swiss National Science Foundation under Award Numbers 10003518 and 237378, and is part of the SYNTHIA project. SYNTHIA (Synthetic Data Generation framework for integrated validation of use cases and AI healthcare applications) is supported by the Innovative Health Initiative Joint Undertaking (IHI JU) under grant agreement No. 101172872. Thus, the project is partially funded by the European Union, the private members, and those contributing partners of the IHI JU. Views and opinions expressed are however those of the authors only and do not necessarily reflect those of the aforementioned parties. Neither of the aforementioned parties can be held responsible for them.

\bibliography{synthetic-uses}
\bibliographystyle{abbrvnat}
\clearpage

\end{document}